\documentclass[letterpaper, 10 pt, journal, twoside]{ieeetran}      

\IEEEoverridecommandlockouts                              

\usepackage{graphicx}
\usepackage{amsmath}
\usepackage{amssymb}
\usepackage{booktabs}
\usepackage{multirow}
\usepackage{makecell}
\usepackage{bm}
\usepackage{enumerate}
\usepackage{arydshln}
\usepackage{booktabs}

\usepackage{tabularx}
\usepackage{algorithmic}
\usepackage{algorithm}

\usepackage{color}
\usepackage{xfrac}
\usepackage{threeparttable}
\usepackage{siunitx}
\usepackage{hhline}

\usepackage{balance}

\usepackage{macros}

\title{
Sparse Semantic Map-Based Monocular Localization in Traffic Scenes Using Learned 2D-3D Point-Line Correspondences
}

\author{Xingyu Chen$^{1}$, Jianru Xue$^{1,\dagger}$, and Shanmin Pang$^{1}$
\thanks{Manuscript received: May, 19, 2022; Revised August, 4, 2022; Accepted September, 1, 2022.}
\thanks{This paper was recommended for publication by Editor Sven Behnke upon evaluation of the Associate Editor and Reviewers' comments.
This work is partially supported by NSFC Projects 62036008 and Development Program of Shaanxi under Grant 2020GY-002.} 
\thanks{$^{1}$The authors are with the Xi'an Jiaotong University. Xi'an, P.R. China.}%
\thanks{$^{\dagger}$Corresponding author's email:
        {\tt\footnotesize jrxue@mail.xjtu.edu.cn}}%
\thanks{Digital Object Identifier (DOI): see top of this page.}
}

\begin{document}

\maketitle

\begin{abstract}

Vision-based localization in a prior map is of crucial importance for autonomous vehicles. Given a query image, the goal is to estimate the camera pose corresponding to the prior map, and the key is the registration problem of camera images within the map.
While autonomous vehicles drive on the road under occlusion (e.g., car, bus, truck) and changing environment appearance (e.g., illumination changes, seasonal variation),
existing approaches rely heavily on dense point descriptors at the feature level to solve the registration problem, entangling features with appearance and occlusion. As a result, they often fail to estimate the correct poses. To address these issues, we propose a sparse semantic map-based monocular localization 
method, which solves 2D-3D registration via a well-designed deep neural network.
Given a sparse semantic map that consists of simplified elements (e.g., pole lines, traffic sign midpoints) with multiple semantic labels, the camera pose is then estimated by learning the corresponding features between the 2D semantic elements from the image and the 3D elements from the sparse semantic map.
The proposed sparse semantic map-based localization approach is robust against occlusion and long-term appearance changes in the environments. 
Extensive experimental results show that the proposed method 
outperforms the state-of-the-art approaches.

\end{abstract}


\section{INTRODUCTION}

\IEEEPARstart{T}{here}
has been an increasing demand for autonomous vehicles in recent years. To achieve autonomous capability, autonomous vehicles need to locate in a prior map. Due to the low-cost, high resolution, and rich color of camera images, monocular localization has received significant attention. Given a query image, the goal is to estimate the camera pose corresponding to the prior map,
the key problem is the registration of camera images within the map.
Most existing work on map-based localization  assumes that correspondences are established between the 2D query image and the database of geo-tagged 2D images \cite{valgren2010sift,ulrich2000appearance} or 3D map \cite{sattler2016efficient, li2012worldwide}, by matching local image features (e.g., SIFT \cite{lowe1999object}, ORB \cite{rublee2011orb}). 

Unfortunately, in many scenarios, the query images and the images used to create the map are captured under different appearances, or even under occlusion. For example, illumination changes and seasonal variation lead to appearance changes, and dynamic objects contribute towards occlusions, such as cars, buses, and trucks around robots. As a result, monocular localization under different visual conditions using the aforementioned methods often fail to estimate the correct pose of the camera \cite{sattler2018benchmarking}.

To resolve these challenging issues, the idea of visual localization with a semantic map is proposed~\cite{qin2020avp,qin2021light}. The semantic map is employed for 
solving the registration problem,
which is compact for storage and robust to occlusion and long-term changes in appearance.
The semantic map in these works contains the graphs on the ground plane, the localization problem is then formulated into a semantic ICP problem by inverse perspective mapping the points on the 2D image plane to the 3D ground plane. However, these methods work for lane lines on the ground but can not been used for traffic signs or poles in the air.


Conventional Perspective-n-Points (PnP) methods are viable to solve the 2D-3D registration problem when the correspondences are given by matching query image features and map features. However, while using a sparse semantic map that is made up of simplified and standardized elements (e.g., lines of poles, midpoints of traffic signs), the correspondences between 2D and 3D elements are often not known a priori, the task becomes non-trivial chicken-and-egg problem since the estimation of correspondence and pose is coupled.

In this paper, we simultaneously solve for both the 6-DoF camera pose and 2D-3D correspondences by the proposed blind PnP method. Firstly, we integrate off-the-shelf convolutional neural networks 
to detect standardized 2D semantic elements and extract the discriminative matching features of 2D and 3D semantic elements by a two-stream neural network. The learned features encode local geometric structure, global context, and semantic topologies of elements, which do not rely on the features of raw images. Secondly, an optimal transport based global feature matching module is employed to estimate a joint correspondence probability matrix among all 2D-3D pairs. Sorting the 2D-3D matches in decreasing order with their probabilities produces a prioritized match list. Thirdly, we exploit the obtained prioritized match list in the P3P-RANSAC approach \cite{gao2003complete,fischler1981random} and optimize the joint correspondence residuals in a weighted Perspective-n-Points-and-Lines (PnPL) manner.
The contribution of this paper is summarized as follows:
\begin{enumerate}
\item 
An efficient learning based blind PnP approach for 2D-3D registration is proposed for monocular localization using a sparse semantic map, which integrates correspondence learning with the PnP algorithm.

\item A robust correspondence learning module is developed to extract the discriminative matching features of 2D and 3D semantic elements and estimate the joint correspondence probability matrix.

\item A differentiable weighted PnPL module is modeled point-line-wise to optimize correspondence residuals.
\end{enumerate}

\begin{figure*}[t]
\centering
  \includegraphics[width=0.8\linewidth]{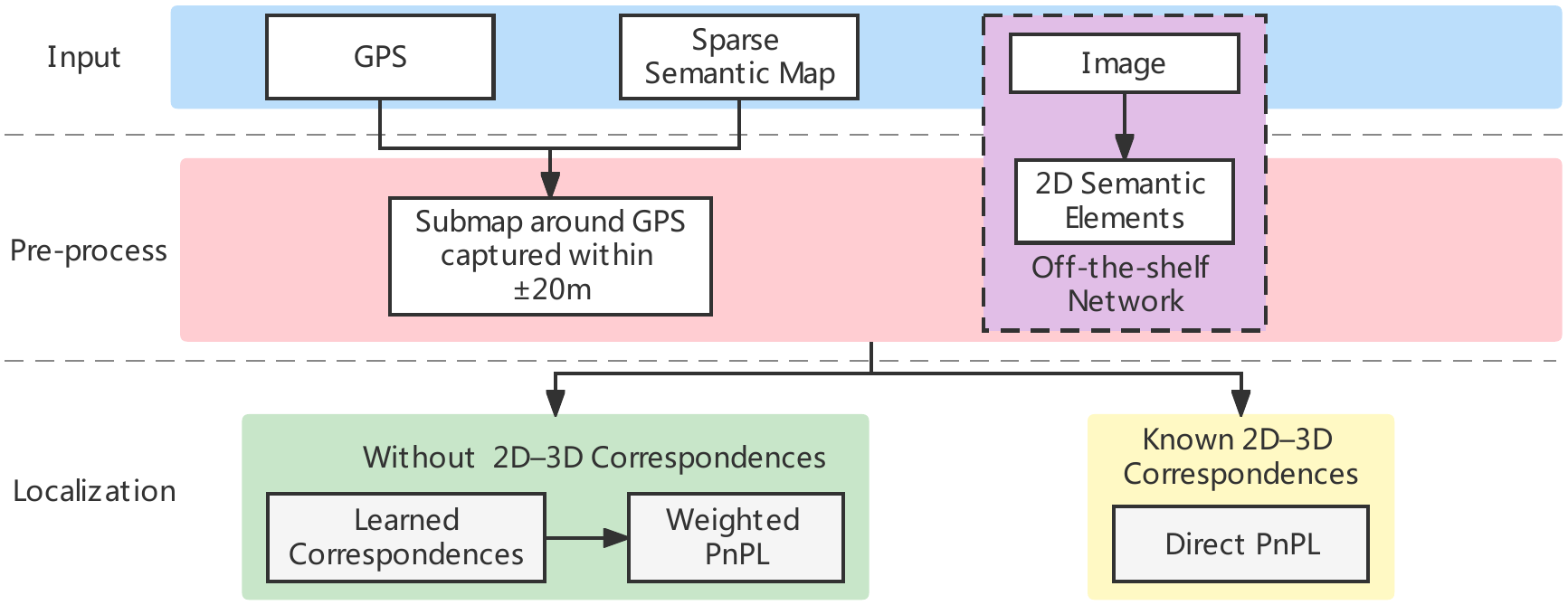}
   \caption{Overview of the proposed sparse semantic map-based localization pipeline. The blue blocks are inputs of our system, where we assume the sparse semantic map is built offline. The red blocks show the pre-processing steps to get the standardized 2D and 3D semantic elements. The yellow block indicates that we perform direct PnPL if the scene is simple that correspondences are directly given by semantic information. Most of the time, the search space of correspondences is enormous, and outliers are prevalent. The green blocks indicate the solution for complex scenes.}
\label{fig:pipeline}
\end{figure*}

\section{RELATED WORK}

A traditional map-based localization pipeline usually consists of feature extraction, feature matching, and optimization. Given different types of queries and maps, there are various methods to implement the pipeline. Here, we focus on the issues of feature matching and optimization, and divide these localization technologies into three categories: 2D-2D, 3D-3D, and 2D-3D.

\subsection{2D-2D}

2D-2D localization \cite{xu2020probabilistic,clement2020learning} is usually the first step in current Simultaneous Localization and Mapping (SLAM) \cite{mur2015orb} or Structure from Motion (SfM) \cite{schoenberger2016sfm}. It is done in the image space. Given a query image, the goal is to retrieve the image corresponding to the same place from a database of geo-tagged images and estimate the query pose via the retrieved image. Typically 2D-2D localization includes feature extraction and matching, optionally followed by a relative pose estimation solver. First, a set of local features such as SIFT \cite{lowe1999object} or ORB \cite{rublee2011orb} are extracted from both query and database images. Then the local features of the query image are aggregated to a global descriptor
\cite{valgren2010sift,ulrich2000appearance} and used to approximate the pose of the query image via matching against its nearest neighbor in database images. Dense correspondences can be established based on the local features, which is used to solve for the finer pose using the five-point algorithm \cite{nister2004efficient}, or eight-point algorithm \cite{longuet1981computer}, etc. However, these 2D-2D localization methods often failed under different visual condition. Such as illumination changes, weather, season, and occlusion \cite{sattler2018benchmarking}.


\subsection{3D-3D}

The availability of 3D information such as Lidar enables 3D-3D localization \cite{miller2021any}. Existing work can be classified into two categories: (1) blind correspondences methods and (2) two-step feature-based methods. Methods like ICP \cite{besl1992method}, NDT \cite{biber2003normal}, and Go-ICP \cite{yang2013go} can estimate the query pose without establishing correspondences between the 3D observation and the prior 3D map. Two-step feature-based methods use 3D feature detectors \cite{tombari2013performance,li2019usip} and descriptors \cite{tombari2010unique,gojcic2019perfect} to establish the correspondences by matching the features, then estimates pose using the Singular Value Decomposition (SVD) solver \cite{golub1971singular}. The 3D-3D localization methods are widely used in Lidar-based SLAM algorithms like LOAM \cite{zhang2014loam}, Cartographer \cite{hess2016real}, etc. However, the expensive Lidar and memory-consuming dense 3D map make it unaffordable for the mass production of mobile robots.

\subsection{2D-3D}

Most existing work on 2D-3D localization assumes that correspondences are first established from local image features of the 2D query image and the 3D map \cite{sattler2016efficient, li2012worldwide}. The 3D map is usually built from a collection of images using SfM \cite{schoenberger2016sfm}, and the associated local features, e.g., SIFT \cite{lowe1999object}, are stored with the map. To estimate the pose, each 2D-3D feature pair votes for its correspondence independently, without considering other pairs in the image. Then a minimal solver algorithm, e.g., EPnP \cite{lepetit2009epnp}, combined with RANSAC \cite{fischler1981random} iterations, is used for pose estimation. Similar to 2D-2D localization, these approaches require local image features that suffers from domain shifts. 
To close the domain gap, methods \cite{feng20192d3d,li2021deepi2p} are proposed, which employ neural networks to extract domain-invariant features from 2D images and dense 3D maps. Nevertheless, these methods are still hampered by the difficulty of extracting domain-invariant features from appearance-variant images. The BPnPNet \cite{campbell2020solving} is designed to extract domain-invariant features from 2D-3D points and solve the point-wise PnP problem. However, it suffers from noise points and clutter maps. In contrast, our method tries to learn the features of sparse 2D and 3D semantic elements that are simplified and standardized, which means it is robust across all domains. And we solve the PnP problem in the point-line-wise, which takes more advantage of the geometry constraints than point-wise PnP.




\section{Overview}
\label{sec:Method}

\begin{figure*}[!t]
\centering
  \includegraphics[width=0.8\linewidth]{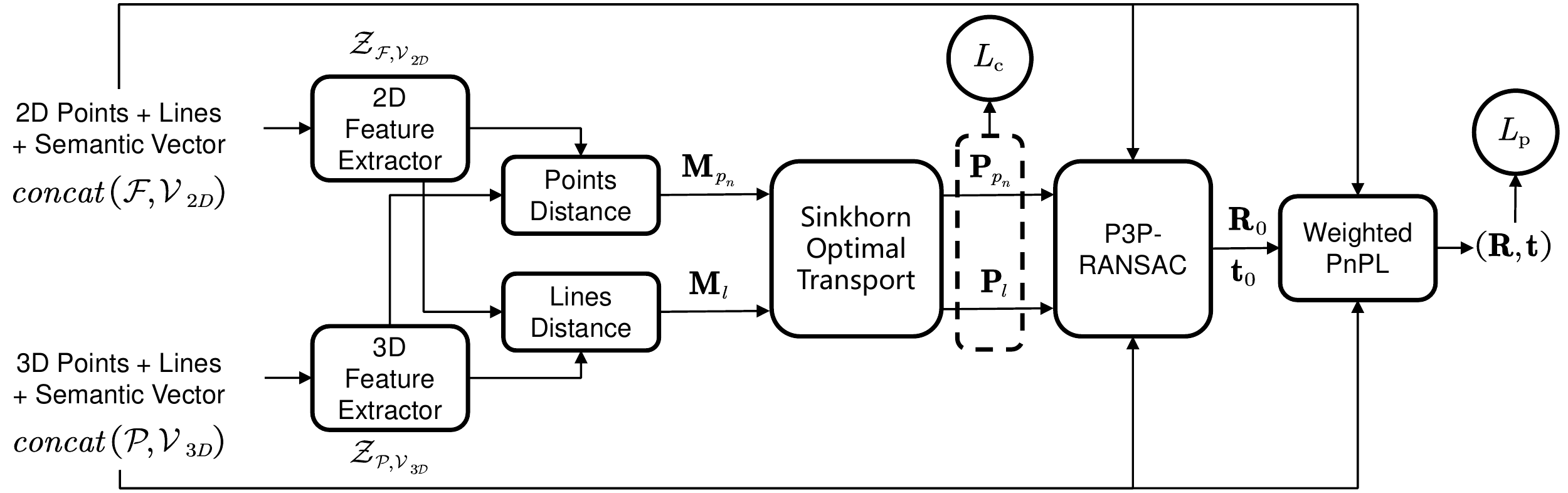}
   \caption{Framework of our pose estimator. The input is a set of 2D and 3D elements coordinates, direction, and semantic vector, from which point-wise discriminative matching features are extracted using a two-stream network. Feature matching is then performed by computing the L2 distance between the 2D and 3D features and using the Sinkhorn algorithm to estimate a joint probability matrix among all 2D-3D pairs. To estimate the camera pose, We exploit the obtained joint correspondence probability matrix in the P3P-RANSAC approach and optimize correspondence residuals in a weighted Perspective-n-Points-and-Lines (PnPL) manner.}
\label{fig:framework}
\end{figure*}

We propose to simultaneously estimate both the 6-DoF camera pose and sparse semantic 2D-3D correspondences, as shown in Fig.~\ref{fig:pipeline}. Specifically, we propose to build sparse semantic maps based on standardized semantic elements offline and use such maps for online localization. Given the GPS, we first get the initial guess of the pose with the error of 10 meters or so. We use GPS as the origin and a radius of 20 meters to crop a submap. In order to find the correspondences between the camera image and the submap, we integrate off-the-shelf deep models to detect standardized 2D elements. If the scene is simple, the correspondences could be obtained directly by semantic information, we will perform the direct PnPL algorithm to get the camera pose. In many cases, the search space of correspondences is enormous, and outliers are prevalent, so we propose to learn correspondences by a neural network, and optimize correspondence residuals in a weighted PnPL manner.

\section{Correspondence Learning}
\subsection{2D semantic Detection and Representation}
\label{sec:Elements}

The involved semantic elements include two major types: lines of poles and midpoints of traffic signs. Given an image, we use YOLOv5 \cite{glenn_jocher_2022_6222936} to detect the bounding box of the 2D elements, then we expand the bounding box by $10 \%$ and feed it to deeplabv3+ \cite{chen2018encoder} to segment the 2D elements. We use Canny \cite{canny1986computational} followed by dilation and erosion to get the edge of the 2D elements, and then we parameterize a traffic sign using a midpoint and parameterize a pole using a peak point with a unit vector pointing to the bottom point.

\subsection{Feature Extractor}
\label{sec:Features}

Let $\bp \in \reals^3$ denote a 3D point, and $\bbf \in \reals^3$ denote a unit bearing vector corresponding to a 2D point in the image plane of a calibrated camera.
That is, $\|\bbf\| = 1$ and $\bbf \propto \bK^{-1}[u, v, 1]\transpose$, where $(u, v, 1)$ are the 2D image coordinates in homograph form, and $\bK$ is the known intrinsic camera matrix that comes from pre-calibration of camera. Given a set of bearing vectors $\cF = \{\bbf_i\}_{i=1}^{m}$ detected by semantic element detector and a set of 3D points $\cP = \{\bp_i\}_{i=1}^{n}$ of the corresponding submap, we concatenate them with vectors $\mathcal{V}_{3D}$ and $\mathcal{V}_{2D}$, which describes the direction and semantic information of 3D and 2D point sets, respectively. (For poles, we know a priori that bottom points are easily occluded by moving objects, while peak points are not, so we use peak points, the direction vector pointing to the bottom points, and one hot semantic vector (e.g., $0001$) to represent poles. Accordingly, we use the center point, zero direction vector, and one hot semantic vector (e.g., $0010 \cdots 1000$) to represent traffic signs.) We learn the discriminative features using PointNet \cite{qi2017pointnet} inspired neural networks $\Phi _{\phi}$ and $\Psi _{\psi}$, in which parameters $\Phi$ and $\psi$ account for the local geometric structure, global context, and semantic topologies at each element of the 2D and 3D input, respectively:
\begin{equation}\small
\label{eq1}
\begin{split}\small
\mathcal{Z}_{\mathcal{F}} &= \Phi _{\phi}\left( concat\left( \mathcal{F},\mathcal{V}_{2D} \right) \right), \\
\mathcal{Z}_{\mathcal{P}} &= \Psi _{\psi}\left( concat\left( \mathcal{P},\mathcal{V}_{3D} \right) \right),
\end{split}
\end{equation}
where $\mathcal{Z}_{\mathcal{F}}=\left\{ \mathbf{z}_{\mathbf{f}_i} \right\} _{i=1}^{m}$, $\mathcal{Z}_{\mathcal{P}}=\left\{ \mathbf{z}_{\mathbf{p}_i} \right\} _{i=1}^{n}$ and the latent vector $\mathbf{z}_{\mathbf{f}_i}\in \mathbb{R}^{128}$, $\mathbf{z}_{\mathbf{p}_i}\in \mathbb{R}^{128}$ are the discriminative features of 2D and 3D respectively. 

To extract point-wise features, similar to Liu \etal \cite{liu2020learning}, we use a point-wise KNN graph to aggregate features. Firstly, we build a point-wise KNN graph and concatenate the anchor point and edges (residuals features of anchor point and neighbor points), then we perform multilayer perceptron (MLP) to extract the features followed by average pooling. We replicate the process $12$ times with residual connections \cite{he2016deep}, and extract features from local to global with the help of the KNN graph, where the feature of different semantic elements will not aggregate together in the first few layers of the network since there is a one hot semantic distance between them.

\subsection{Feature Matcher}
Given the learned features $\mathcal{Z}_{\mathcal{F}}$, $\mathcal{Z}_{\mathcal{P}}$, we perform feature matching to estimate the likelihood of a given 2D-3D pair. To do so, we use the pairwise L2 distance in Euclidean space to establish the distance matrix $\mathbf{M}\in \mathbb{R}_{+}^{m\times n}$ between 2D and 3D features, which measures the cost of assigning 2D elements to 3D elements:
\begin{equation}\small
\label{eq2}
\mathbf{M}_{ij}=\lVert \mathbf{z}_{\mathbf{f}_i}-\mathbf{z}_{\mathbf{p}_j} \rVert _2.
\end{equation}

Furthermore, to globally handle the likelihood of pairwise match, we describe the matching ambiguities as an optimal transport problem \cite{villani2009optimal} and solve the joint correspondence probability matrix $\mathbf{P}$ by solving:
\begin{equation}\small
\label{eq3}
\underset{\mathbf{P}\in U\left( \mathbf{r,s} \right)}{\arg \min }\sum_{i=1}^m{\sum_{j=1}^n{\left( \mathbf{M}_{ij}\mathbf{P}_{ij}+\mu \mathbf{P}_{ij}\left( \log \mathbf{P}_{ij}-1 \right) \right)}},
\end{equation}
where $U\left( \mathbf{r,s} \right)$ is the transport polytope that couples two prior probability vectors $\mathbf{r}$ and $\mathbf{s}$, given by:
\begin{equation}\small
\label{eq4}
U\left( \mathbf{r,s} \right) =\left\{ \mathbf{P}\in \mathbb{R}_{+}^{m\times n}\mid \mathbf{P1}^n=\mathbf{r,P}^{\top}1^m=\mathbf{s} \right\}.
\end{equation}

The prior probability vectors $\mathbf{r} \in \mathbb{R}_{+}^{m}$ and $\mathbf{s} \in \mathbb{R}_{+}^{n}$ with $\sum \mathbf{r}=1$ and $\sum \mathbf{s}=1$, which represent the likelihood that a 2D or 3D element has a valid match and is not an outlier. In this work, we use the uniform priors $\mathbf{r}=\frac{1}{m} \mathbf{1}$ and $\mathbf{s}=\frac{1}{n} \mathbf{1}$, as introduced by Campbell \etal \cite{campbell2020solving}, which means that each element has the same prior likelihood of matching.

We use the Sinkhorn algorithm \cite{marshall1968scaling} to solve the optimal transport problem Eq.~\ref{eq3}, as has been previously demonstrated in the literature \cite{campbell2020solving}.

More precisely, different semantic elements will pass through different branches of the feature matcher. Each one is used to optimize the correspondence probability matrix $\mathbf{P}_{p_1} \cdots \mathbf{P}_{p_n}$ or $\mathbf{P}_{l}$ of the specific semantic elements, given the distance matrix $\mathbf{M}_{p_1} \cdots \mathbf{M}_{p_n}$ or $\mathbf{M}_{l}$ of specific traffic signs and poles, respectively, as illustrate in Fig.~\ref{fig:framework}

\section{Pose Estimation}
\subsection{Coarse Pose Estimator}
The objective of the coarse pose estimator is to find the rotation $\bR \in SO(3)$ and translation $\bt \in \reals^3$ that transforms $\cP$ to the coordinate system of $\cF$ with the greatest number of inlier correspondences. Given the correspondences of the peak points of the poles, we can solve the correspondences of the bottom points at the same time. Although the outlier correspondences of the bottom points are easy to be introduced due to the occlusion, we can utilize the correspondences of bottom points to solve the pose with the help of the RANSAC \cite{fischler1981random} algorithm, which is robust to outliers.

Specifically, we solved the pose through the P3P \cite{gao2003complete} algorithm with RANSAC, which repeatedly selects 3 pairs of matching elements in a subset of prioritized matches and 1 pair of verification points to estimate the pose and then chooses the solution with the highest inlier rate as the coarse pose estimation. To obtain the subset of prioritized matches, we reshape $\mathbf{P}$ into a correspondence probability vector, sort it by decreasing probability and truncate it to obtain the Top-K prioritized matching subset with high correspondence probability. The indicator that correspondence is inlier is defined by an angular threshold $\theta$ of the reprojection error:
\begin{equation}\small
\label{eq5}
\arccos \left(
\frac{\mathbf{f}_{i}^{\top} (\mathbf{R} \mathbf{p}_{j}+\mathbf{t})}{\left\|\mathbf{R} \mathbf{p}_{j}+\mathbf{t}\right\|}\right),
\end{equation}
where $\mathbf{p}_{j}$ is the closest point of $\cP$ reprojected to the coordinate of $\cF$, with respect to $\mathbf{f}_{i}$, given the estimated rotation $\bR$ and translation $\bt$. The solution $\mathbf{R}_0$ and $\mathbf{t}_0$ of the robust randomized global search is used as a coarse pose to initialize the weighted PnPL optimization.

\subsection{Weighted PnPL}
We now have a coarse pose, a correspondence probability matrix $\mathbf{P}_{p}$ of points, a correspondence probability matrix $\mathbf{P}_{l}$ of lines, then we refine the camera pose by optimizing:
\begin{equation}\small
\label{eq6}
\underset{\mathbf{r} \in \mathbb{R}^{3},\mathbf{t} \in \mathbb{R}^{3}}{\arg \min }f_p(\mathbf{P}_{p}, \mathbf{r}, \mathbf{t})+f_l(\mathbf{P}_{l}, \mathbf{r}, \mathbf{t}),
\end{equation}
over $\br$ and $\bt$, where $\br$ is an angle-axis representation of the rotation such that $\bR_{\br} = \exp{[\br]_{\times}}$. The operator $[\cdot]_{\times}$ is the skew-symmetric operation, and the closed-form exponential map constrains $\bR \in SO(3)$. The $f_p$ and $f_l$ are the objective functions for points and lines, the specific formulation is:
\begin{equation}\footnotesize
\label{eq7}
\begin{split}
&f_p(\mathbf{P}_{p}, \mathbf{r}, \mathbf{t}) = \sum_{i=1}^{m} \sum_{j=1}^{n} \mathbf{P}_{p i j}\left(1-\mathbf{f}_{i}^{\top} \frac{\mathbf{R}_{\mathbf{r}} \mathbf{p}_{j}+\mathbf{t}}{\left\|\mathbf{R}_{\mathbf{r}} \mathbf{p}_{j}+\mathbf{t}\right\|}\right), \\
&f_l(\mathbf{P}_{l}, \mathbf{r}, \mathbf{t})= \\
&\sum_{i=1}^{m2} \sum_{j=1}^{n2} \mathbf{P}_{l i j}\left(1-\left\|\frac{\mathbf{f}_{i} \times \mathbf{v}_{2di}}{\left\|\mathbf{f}_{i} \times \mathbf{v}_{2di}\right\|} \times \frac{\mathbf{R}_{\mathbf{r}} (\mathbf{p}_{j}+\mathbf{v}_{3dj})+\mathbf{t}}{\left\|\mathbf{R}_{\mathbf{r}} (\mathbf{p}_{j}+\mathbf{v}_{3dj})+\mathbf{t}\right\|} \right\|\right),
\end{split}
\end{equation}
where $\mathbf{v}_{2d}$, $\mathbf{v}_{3d}$ is the unit direction vector of lines in 2D and 3D respectively. 
On the one hand, a point in the image corresponds to a ray in space and $f_p$ encourages the transformed point $\mathbf{R}_{\mathbf{r}} \mathbf{p}_{j}+\mathbf{t}$ to lie on the ray spanned by $\mathbf{f}_{i}$, i.e. $\mathbf{R}_{\mathbf{r}} \mathbf{p}_{j}+\mathbf{t}$ and $\mathbf{f}_{i}$ be collinear. On the other hand, a line in the image corresponds to a plane in space and $f_l$ encourages the transformed point $\mathbf{p}_{j}+\mathbf{v}_{3dj}$ to lie in the plane spanned by $\mathbf{f}_{i}$ and $\mathbf{v}_{2di}$, i.e. be perpendicular to $\mathbf{f}_{i} \times \mathbf{v}_{2di}$.

We minimize the full nonlinear function using the L-BFGS optimizer \cite{byrd1995limited} to get the finer $\mathbf{R}$ and $\mathbf{t}$ for the given joint probability matrix.

\subsection{Direct PnPL}
When the scene is simple that the correspondences are directly given by the semantic information, we will perform the direct PnPL:
\begin{equation}\small
\label{eq8}
\underset{\mathbf{r} \in \mathbb{R}^{3},\mathbf{t} \in \mathbb{R}^{3}}{\arg \min }f_p(\mathbf{P}^{D}_{p}, \mathbf{r}, \mathbf{t})+f_l(\mathbf{P}^{D}_{l}, \mathbf{r}, \mathbf{t})
\end{equation}
where the $\mathbf{P}^{D}_{p}$ and $\mathbf{P}^{D}_{l}$ is the Boolean one-to-one correspondence matrix with at most one nonzero element in each row and column. Most of the time, we perform weighted PnPL because the search space of correspondences is enormous, and outliers are prevalent. 

\section{The Joint Learning Method}
To optimize the parameters $\Phi$ and $\psi$ of the feature extractor, we minimize two loss functions. The first is a correspondence loss $L_{\mathrm{c}}$ to bring the estimated correspondence matrix $\mathbf{P}_{p}$ closer to the ground truth:
\begin{equation}\small
\label{eq9}
L_{\mathrm{c}}=\sum_{i}^{m} \sum_{i}^{n}\left(1-2 \mathbf{C}_{i j}^{\mathrm{gt}}\right) \mathbf{P}_{i j}, \mathbf{t}),
\end{equation}
where the ground-truth correspondence matrix $\mathbf{C}_{i j}$ is $1$ if $\left\{\mathbf{f}_{i}, \mathbf{p}_{j}\right\}$ is a true correspondence and $0$ otherwise. Optimizing the $L_{\mathrm{c}}$ encourages the feature extractor $\Phi _{\phi}$ and $\Psi _{\psi}$ to maximize the joint probability of inlier correspondences and minimize the joint probability of outlier correspondences.

The second is a pose loss $L_{\mathrm{p}}$ to encourage the network to generate correspondence matrices that are amenable to the weighted PnPL solver:
\begin{equation}\small
\label{eq10}
L_{\mathrm{p}} =arccos \frac{1}{2}\left(\operatorname{trace} \mathbf{R}_{\mathrm{gt}}^{\top} \mathbf{R}-1\right) + \left\|\mathbf{t}-\mathbf{t}_{\mathrm{gt}}\right\|_{2}.
\end{equation}

To achieve accurate and robust pose estimation, we combine the aforementioned loss and jointly train all the parameters to optimize the full objective:
\begin{equation}\small\label{eq11}
L=L_{\mathrm{c}}+\gamma_{\mathrm{p}} L_{\mathrm{p}}.
\end{equation}
Where the first term is balanced by the second, which corresponds to a regularizer with a multiplier $\gamma_{\mathrm{p}}$, and this discourages the feature extractor from ignoring the demand of the pose estimator. Note that loss functions $L_{\mathrm{c}}$ and $L_{\mathrm{p}}$ are complementary while generating a perfect correspondence matrix is not achievable in practice. There are a series of suboptimal corresponding matrices, which achieves the same value of $L_{\mathrm{c}}$ but yields distinct value of $L_{\mathrm{p}}$. The gradients backward from the pose loss to the feature extractor guarantees that the weighted PnPL pose estimator indeed benefits from the extracted features and correspondences.

\section{EXPERIMENTS}

\subsection{Implementation Details}
We closely follow the procedure of YOLOv5 \cite{glenn_jocher_2022_6222936} and deeplabv3+ \cite{chen2018encoder} to train the element detector. The rest of our framework is implemented in PyTorch follows \cite{liu2020learning,li2021deepi2p,campbell2020solving}. We train the feature extractor using the Adam optimizer \cite{kingma2014adam} with a learning rate of $5\times10^{-4}$. We use a batch size of $12$ and train for $120$ epochs with the correspondence loss only ($\gamma_{\mathrm{p}}=0$), followed by $80$ epochs with the pose loss as well ($\gamma_{\mathrm{p}}=1$), because the pose loss works only when the correspondence probability matrix $\mathbf{P}_{p}$ has been warm started. For the KNN-graph of the feature extractor, we set the number of neighbors to $4$. The entropy parameter $\mu$ of the Sinkhorn algorithm is set to $0.1$. The inlier reprojection threshold and the maximum number of iterations of P3P-RANSAC are set to $0.003$ and $1000$. 
The back-propagation of gradients through the optimization layers (weighted PnPL and Sinkhorn) is interpreted as a bi-level optimization problem and solved by the implicit differentiation technique \cite{gould2021deep}. All experiments are run on a single Titan Xp GPU.











\begin{table*}[t]
\caption{Localization Accuracy on the KITTI Datasets}
\vspace{-0.2cm}
\label{table_example}
\begin{center}
\small
\begin{tabular}{ll|cccc|cccc}
\hline
                                                                   &                                                     & \multicolumn{4}{c|}{RTE $(\mathrm{m})$}                       & \multicolumn{4}{c}{RRE $\left({ }^{\circ}\right)$}            \\ \cline{3-10} 
                                                                   &                                                     & Mean          & Q1            & Q2            & Q3            & Mean          & Q1            & Q2            & Q3            \\ \hline
\multicolumn{1}{l|}{\multirow{2}{*}{Running on Raw Data}}          & 2D3D-MatchNet \cite{feng20192d3d}  & 27.1          & 12.0          & 23.3          & 63.1          & 134           & 115           & 127           & 162           \\
\multicolumn{1}{l|}{}                                              & DeepI2P \cite{li2021deepi2p}       & 42.9          & 15.4          & 32.7          & 96.2          & 145           & 82.4          & 131           & 167           \\ \hline
\multicolumn{1}{l|}{\multirow{4}{*}{Running on Detected Elements}} & Point 2D3D-MatchNet*                                & 18.4          & 6.72          & 13.5          & 27.3          & 75.8          & 23.1          & 53.7          & 146           \\
\multicolumn{1}{l|}{}                                              & Point DeepI2P*                                      & 3.94          & 0.57          & 1.39          & 5.83          & 24.1          & 5.37          & 13.2          & 36.5          \\
\multicolumn{1}{l|}{}                                              & BPnPNet \cite{campbell2020solving} & 1.14          & 0.21          & 0.49          & 1.57          & 2.10          & 0.52          & 0.74          & 2.69          \\
\multicolumn{1}{l|}{}                                              & Ours w/o Semantics                       & 0.49          & 0.14          & 0.21          & 0.97          & 0.92          & 0.17          & 0.35          & 1.83          \\ \hline
\multicolumn{1}{l|}{\multirow{4}{*}{Running on Semantic Elements}} & Semantic Point 2D3D-MatchNet*                       & 5.21          & 2.14          & 4.32          & 12.6          & 24.3          & 6.49          & 18.6          & 37.2          \\
\multicolumn{1}{l|}{}                                              & Semantic Point DeepI2P*                             & 2.17          & 0.25          & 0.71          & 3.24          & 8.69          & 2.15          & 6.21          & 12.4          \\
\multicolumn{1}{l|}{}                                              & Semantic BPnPNet*                                   & 0.36          & 0.17          & 0.25          & 0.82          & 0.65          & 0.23          & 0.39          & 1.22          \\
\multicolumn{1}{l|}{}                                              & Ours                                                & \textbf{0.22} & \textbf{0.09} & \textbf{0.18} & \textbf{0.29} & \textbf{0.34} & \textbf{0.14} & \textbf{0.26} & \textbf{0.45} \\ \hline
\end{tabular}
\begin{tablenotes}
        \centering
        * Methods are modified by us to utilize the standardized elements and semantic information.
\end{tablenotes}
\vspace{-0.6cm}
\end{center}
\label{tab:error}
\end{table*}

\subsection{Evaluation}

We evaluate our method 
on the KITTI \cite{geiger2013vision} dataset with a second construction. We follow the common practice of utilizing the $0$-$8$ sequences for training and $9$-$10$ for testing.

\noindent\textbf{Evaluation of Element detector.}
For the element detector, we select one image per 10 frames of the training sequence to train the object detection network YOLOv5. We label the bounding box for 4098 images with four types of traffic elements, including poles, triangular, rectangular, and rounded signs. Then, we use the bounding box to crop images to label the semantic information at the pixel level. Combining them with the data that we have cropped and labeled on TT100k \cite{zhu2016traffic}, we use a total of $1157$ images with the size of $128 \times 128$ to train the Deeplabv3+ network.

\begin{figure*}[t]
\centering
  \includegraphics[width=0.8\linewidth]{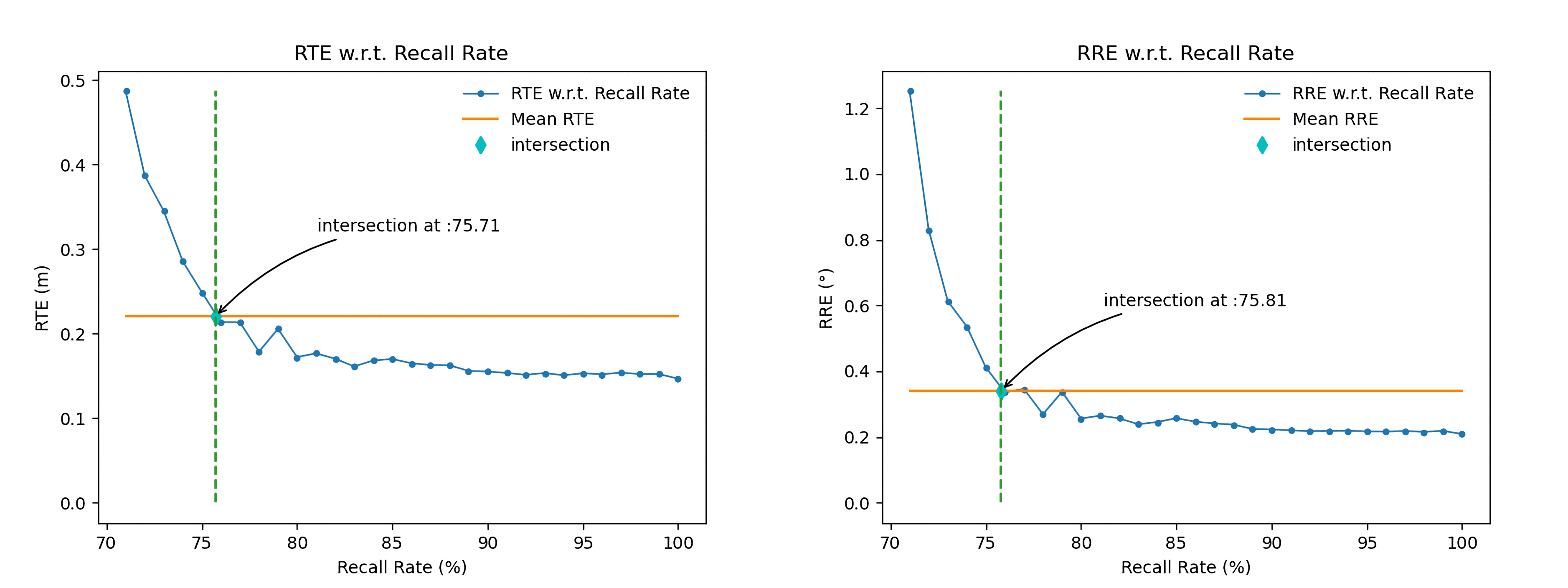}
   \caption{Analysis of RTE and RRE with respect to Recall Rate. x-axis is the Recall Rate of a single image, and y-axis is the statistical RTE and RRE when the element detector is under that specified Recall Rate.}
  \vspace{-0.4cm}
\label{fig:recal}
\end{figure*}

We evaluate the performance of the element detector on $216$ images with three criteria: Recall Rate (RR), Root Mean Square Error (RMSE), Mean of Error (ME), and Standard Deviation of Error (SDE). As shown in Table \ref{tab:detec}, the RR of our element detector is about $(80\%)$, which is sufficient for the pose estimation since our weighted PnPL is robust for missing detection, as shown in Fig.~\ref{fig:recal}. It indicates the statistical Relative Translation Error (RTE) and Relative Rotation Error (RRE) when the element detector has different Recall Rate. 
Furthermore, we report the RMSE, ME, and SDE of our, which is acceptable for pose estimation when the image resolution is $1382 \times 512$.

\noindent\textbf{Evaluation of Localization Accuracy.}
To construct the data for training the feature extractor and evaluate the localization accuracy, we need to build the 3D elements map. 
Considering that it is difficult for horizontal Lidar to capture the high poles and high traffic signs, we need to build a 3D map through images.
Firstly, We label and match the 2D elements between the left and right view images of the KITTI data set. Secondly, we use the linear triangulation method to reconstruct the 3D elements based on the matched 2D elements. Finally, we utilize the optimized global pose \cite{behley2021ijrr}
as the ground truth to construct the 3D elements map, followed by the DBSCAN \cite{ester1996density} algorithm for de-duplication. 
Similar to the criteria of element detector, we report the RMSE, ME and SDE reprojection error between the reprojected 2D elements of constructed 3D map and 2D annotation elements in Table \ref{tab:map}.
The reprojection error between the 3D elements map and 2D annotation elements is less than $1 ~\mathrm{pixel} / 1^{\circ}$, indicating that the accuracy of the 3D elements map meets the requirements of localization. The image-submap pairs are sampled within $\pm 20m$. We apply the data augmentation on 3D elements by random rotation within 360 degrees and random translation within $5\mathrm{m}$ on the horizontal plane.
Note that since pose estimation requires at least four 2D elements, only images with 2D elements greater than or equal to 4 are considered valid data, and other images will not be included in the training and test set. The percentage of valid data in all sequences is $3.29\%$.
In total, there are $2984$ frames for training and $459$ frames for testing.

\begin{figure*}[t]
\centering
  \includegraphics[width=0.8\linewidth]{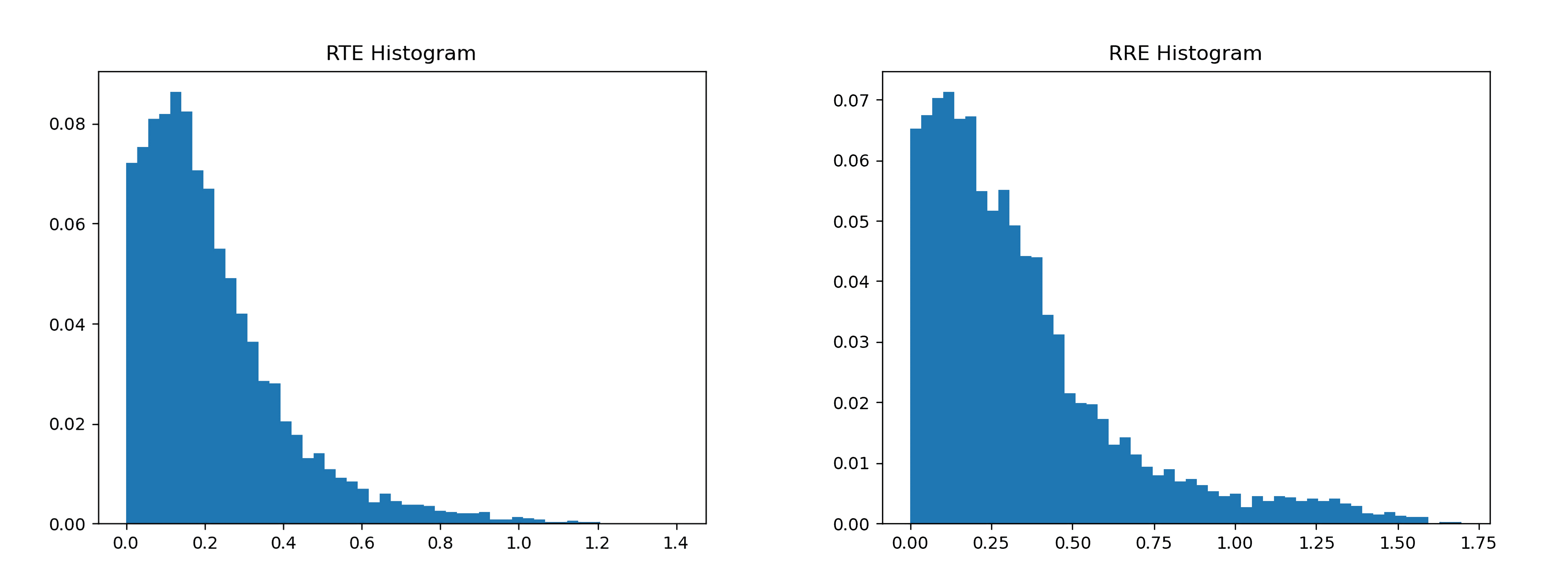}
   \caption{Histograms of localization accuracy RTE and RRE on the KITTI datasets. x-axis is RTE $(\mathrm{m})$ and RRE $\left({ }^{\circ}\right)$, and y-axis is the percentage.}
  \vspace{-0.4cm}
\label{fig:Histogram}
\end{figure*}

Following the practice of \cite{li2021deepi2p}, the localization accuracy is evaluated with two criteria: Relative Translation Error (RTE) and Relative Rotation Error (RRE).
The distribution of the localization accuracy of RTE $(\mathrm{m})$ and RRE $\left({ }^{\circ}\right)$ on the KITTI dataset is shown in Fig.~\ref{fig:Histogram}. Specifically, the average of the translational/rotational errors are $ 0.22 \mathrm{m} / 0.34^{\circ}$, the median of the translational/rotational errors are $ 0.18 \mathrm{m} / 0.26^{\circ}$. We achieve the robust performance where $ 99.5\%$ of the translational errors are less than $1.0 \mathrm{m}$, and $ 94.7 \%$ of the rotational errors are less than $1.0^{\circ}$.

\noindent\textbf{Comparisons.}
We compare our method against BPnPNet \cite{campbell2020solving}, DeepI2P \cite{li2021deepi2p}, 2D3D-MatchNet \cite{feng20192d3d} and five variants of them.
All methods are trained and tested using the same training and test set,
and are compared in terms of the localization accuracy, except DeepI2P and 2D3D-MatchNet, where we use their pre-trained network and fine-tune the parameters in our training set since train a network from scratch to extract the features from the original image requires more data. 
We report the mean and quartiles for translation error RTE $(\mathrm{m})$ and rotation error RRE $\left({ }^{\circ}\right)$ according to baselines. We denote the first, second (median), and third quartiles as Q1, Q2, and Q3, as summarized in Table \ref{tab:error}. 

Both 2D3D-MatchNet \cite{feng20192d3d} and DeepI2P \cite{li2021deepi2p} suffer from the difficulty of extracting the features from raw images and matching the features with 3D elements. To utilize the standardized elements, we modified them to Point 2D3D-MatchNet and Point DeepI2P, which learns to match across 2D elements detected by our element detector and 3D elements of the map. The improvement of the RTE and RRE showed in Table \ref{tab:error} demonstrates that the sparse elements do help with localization accuracy. 

The BPnPNet \cite{campbell2020solving} is designed to solve the blind PnP problem, which is naturally suited to the task of matching 2D-3D elements and pose estimation. It produces a more accurate pose estimation and is able to achieve RTE/RRE error in the single digits. However, the RTE/RRE of BPnPNet still tends to exhibit significant variances, while Ours w/o Semantics does not. Ours w/o Semantics is ablation of our method, which builds upon our full model by eliminating the semantic information from the inputs. 

In order to make full use of the semantic information, we implement the Semantic Point 2D3D-MatchNet and Semantic Point DeepI2P, which have a more accurate pose estimation, but cannot to achieve the accuracy of BPnPNet and Ours w/o Semantics due to the lack of robust differentiable geometric optimization. 

Furthermore, we implement the Semantic BPnPNet to enable the BPnPNet to extract features from both element position and semantic information. However, it still suffers from significant variances in RTE and RRE. These variances are the consequence of Semantic BPnPNet's attempt to split lines into points and use only points to estimate the matching and pose, where the structured information of lines are underutilized and underexploited. Our method exploits benefits of both points and lines and thereby produces better localization accuracy on all metrics than these baselines.

Moreover, we analyze the average runtime for inference. 2D3D-MatchNet and DeepI2P take about 3.1s and 5.7s to extract features from raw data and estimate pose. For BPnPNet and ours, it spends about 0.2s to detect the 2D elements and 0.1s to extract features from elements and achieve the pose estimation.

\begin{table}[t]
\caption{Detection Accuracy on the KITTI Datasets}
\vspace{-0.6cm}
\label{table_example}
\begin{center}
\small
\resizebox{\linewidth}{!}{
\begin{tabular}{l|cccc}
\hline
                                   & Poles & Triangular & Rectangular & Rounded \\ \hline
RR $(\%)$                          & 66    & 89         & 87          & 79      \\
RMSE $(\mathrm{pixel})$              & NA    & 1.81       & 4.10        & 3.17    \\
Lateral ME $(\mathrm{pixel})$      & NA    & -1.38      & -0.80       & -0.61   \\
Lateral SDE $(\mathrm{pixel})$      & NA    & 0.90       & 4.25        & 1.63    \\
Longitudinal ME $(\mathrm{pixel})$ & NA    & -0.86      & -1.74       & 1.49    \\
Longitudinal SDE $(\mathrm{pixel})$ & NA    & 0.84       & 3.43        & 4.31    \\
Angular ME $({ }^{\circ})$         & -1.52 & NA         & NA          & NA      \\
Angular SDE $({ }^{\circ})$         & 1.32  & NA         & NA          & NA      \\ \hline
\end{tabular}
}
\end{center}
\vspace{-0.4cm}
\label{tab:detec}
\end{table}

\begin{table}[t]
\caption{Mapping Accuracy on the KITTI Datasets}
\vspace{-0.6cm}
\label{table_example}
\begin{center}
\small
\resizebox{\linewidth}{!}{
\begin{tabular}{l|cccc}
\hline
                                   & Poles & Triangular & Rectangular & Rounded \\ \hline
RMSE $(\mathrm{pixel})$            & NA    & 0.86       & 0.74        & 0.62    \\
Lateral ME $(\mathrm{pixel})$      & NA    & -0.02      & 0.03        & 0.05    \\
Lateral SDE $(\mathrm{pixel})$      & NA    & 0.83       & 0.76        & 0.46    \\
Longitudinal ME $(\mathrm{pixel})$ & NA    & -0.06      & 0.01        & -0.03   \\
Longitudinal SDE $(\mathrm{pixel})$ & NA    & 0.44       & 0.38        & 0.31    \\
Angular ME $({ }^{\circ})$         & -0.03 & NA         & NA          & NA      \\
Angular SDE $({ }^{\circ})$         & 0.27  & NA         & NA          & NA      \\ \hline
\end{tabular}
}
\end{center}
\vspace{-0.6cm}
\label{tab:map}
\end{table}

\begin{figure}[t]
\centering
  \includegraphics[width=1\linewidth]{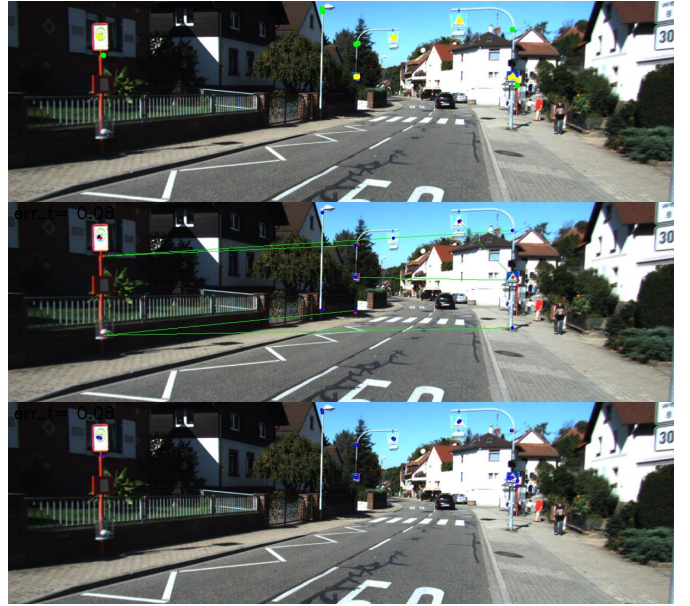}
   \caption{Visualization of the 3D elements map projected onto the images using the ground truth pose (top), where the yellow and green points are the traffic sign and pole elements respectively. Correspondences with outlier (middle), where the blue and red points are the detected and ground truth projected elements, and the green lines between them are correspondences. Reprojection error of pose estimation (bottom), the blue points are the projected 3D elements using the estimated pose, the red points are the projected 3D elements using the ground truth pose, and the green lines between them indicate the reprojection error.
   }
\label{fig:visualization}
\end{figure}

\noindent\textbf{Limitations.} 
As illustrated in Fig.~\ref{fig:visualization}, the 3D elements map suffers from duplication of the same element, which shows that DBSCAN algorithm is not enough for de-duplication. Specific tracking and graph optimization pipelines of standardized elements have to be developed to build a more accurate map. Additionally, although we have built a robust feature matcher, there will still be outliers in the correspondences. Thanks to the P3P-RANSAC and weighted PnPL, we have estimated the correct pose ($\mathrm{RTE}= 0.08 \mathrm{m}$) with tiny reprojection errors. Without exception, given a better feature matcher, the proposed method will achieve better performance in the future.
On the other hand, when 2D semantic elements are less than 4, our method fails. Therefore, how combining vision or inertial odometry with our framework is an essential topic in the future.

\section{CONCLUSION}

Localization in a prior sparse semantic map has grown in prominence and can be utilized in various applications, including mobile robots and autonomous vehicles. While BPnPNet \cite{campbell2020solving} works effectively with 2D and 3D elements, it is hard to achieve low variance pose estimation. 
Experimental results demonstrate that
our method can accurately and robustly locate in a prior sparse semantic map. Specifically, the proposed learning correspondence module can efficiently extract the discriminative matching features of 2D and 3D semantic elements, via simultaneously utilizing the semantic information and element position. Furthermore, the weighted PnPL module optimize the joint correspondence residuals in the point-line-wise, which performs robust differentiable geometric optimization with the points and lines. The advantages of our proposed framework are verified with the constructed KITTI dataset.

{\small
{\bibliographystyle{./IEEEtran} 
\small
\bibliography{./IEEEexample}}
}

\end{document}